\documentclass{article} 
\usepackage{iclr2016_conference,times}
\usepackage{hyperref}
\usepackage{url}
\usepackage{pslatex}
\usepackage{color} 
\usepackage{graphicx}
\usepackage{subfig}
\usepackage{multicol}
\usepackage{multirow}
\usepackage{color}
\usepackage{wrapfig}
\usepackage{longtable}
\usepackage{lipsum}  

\title{Basic Level Categorization Facilitates Visual Object Recognition}

\author{Panqu Wang \\
Department of Electrical and Computer Engineering\\
University of California, San Diego\\
La Jolla, CA 92037, USA \\
\texttt{pawang@ucsd.edu} \\
\And
Garrison W. Cottrell \\
Department of Computer Science \\
University of California, San Diego \\
La Jollca, CA 92037, USA \\
\texttt{gary@ucsd.edu}
}

\begin{document}

\maketitle

\begin{abstract}
Recent advances in deep learning have led to significant progress in the computer vision field, especially for visual object recognition tasks. The features useful for object classification are learned by feed-forward deep convolutional neural networks (CNNs) automatically, and they are shown to be able to predict and decode neural representations in the ventral visual pathway of humans and monkeys. However, despite the huge amount of work on optimizing CNNs, there has not been much research focused on linking CNNs with guiding principles from the human visual cortex. In this work, we propose a network optimization strategy inspired by both of the developmental trajectory of children's visual object recognition capabilities, and \citet{bar2003}, who hypothesized that basic level information is carried in the fast magnocellular pathway through the prefrontal cortex (PFC) and then projected back to inferior temporal cortex (IT), where subordinate level categorization is achieved. We instantiate this idea by training a deep CNN to perform basic level object categorization first, and then train it on subordinate level categorization. We apply this idea to training AlexNet~\citep{krizhevsky2012} on the ILSVRC 2012 dataset and show that the top-5 accuracy increases from $80.13\%$ to $82.14\%$, demonstrating the effectiveness of the method. We also show that subsequent transfer learning on smaller datasets gives superior results.

\end{abstract}

\section{Introduction}
Humans possess the ability to recognize complex objects rapidly and accurately through the ventral visual stream. In a traditional feed-forward view, the ventral visual stream processes the input stimulus from the primary visual cortex (V1), carries the response through V2 and V4, and finally arrives at the interior temporal (IT) cortex, where a more invariant object representation for categorization is obtained~\citep{dicarlo2007}. Among all of the cortical regions, V1 is the best understood, as it can be well-characterized by 2-D Gabor filters~\citep{carandini2005we}, and some subregions in IT are known to be activated by category-specific stimulus, such as faces (FFA and OFA; \citet{kanwisher1997,puce1996differential}), words (VWFA; \citet{mccandliss2003visual}), and scenes (PPA; \citet{epstein1999}). Nevertheless, it remains unclear what the feature representations between V1 and IT are, or how the increasingly complex representations progress through the ventral stream hierarchy, although some answers have been proposed~\citep{cox2014we,gucclu2015deep}.

In the past few years, the advances in deep learning, especially solving computer vision problems using deep convolutional neural networks (CNNs), has shed light on the representations in the ventral visual pathway. Deep CNNs stack computations in a hierarchical way, repeatedly forming 2-D convolutions over the input, applying pooling operation on local regions of the feature maps, and adding non-linearities to the upstream response. By building and training deep CNNs with millions of parameters using millions of images, these systems become the most powerful solutions to many computer vision tasks, such as image classification~\citep{krizhevsky2012,he2015deep}, object detection~\citep{girshick2014rich}, scene recognition~\citep{zhou2014learning}, and video categorization~\citep{karpathy2014large}. Several studies have even shown that these systems are on a par with human performance, in tasks such as image classification~\citep{he2015delving}, and face recognition~\citep{taigman2014deepface}. These results suggest their potential power to help us understand the ventral visual system.

More recently, many studies have been done to optimize and improve the performance of deep CNNs, such as increasing the depth~\citep{Simonyan14c,Szegedy_2015_CVPR}, optimizing the activation function~\citep{he2015delving}, pooling layers ~\citep{lee2015generalizing}, and modifying the loss functions~\citep{lee2014deeply,romero2014fitnets}. Despite the huge success of these engineering approaches, very little has been done to link optimizing deep CNNs by adding guiding principles from the human brain on how visual object recognition is achieved. In fact, while deep CNNs have been used to model and explain the neural data in IT~\citep{yamins2014performance,cadieu2014deep,agrawal2014pixels,guccluetal2014,gucclu2015deep}, inspiration in the opposite direction has not been as evident. 

In this study, we examine the effect of one property of visual object recognition in the brain - the primacy of basic level categorization - as a method for training deep CNNs. This idea is drawn from two different perspectives: behavioral studies of the development of object categorization, and a hypothesized neural mechanism of top-down basic level facilitation in cortex. The basic level is one of three levels of abstraction for categorization of natural objects: subordinate, basic and superordinate. For example, a gala apple (subordinate level) is a type of apple (basic level) which is a type of fruit (superordinate level). Behavioral studies show that the mean reaction times for basic level categorization are fastest~\citep{tanaka1991object}, suggesting the primary role of basic level categories in visual processing. Other studies show that infants and young children categorize at the basic level earlier than the subordinate level~\citep{bornstein2010development}, and even earlier than the superordinate level~\citep{mervis1982order,mandler1988cradle,behl1996basic}. 

In terms of adult visual processing, \citet{bar2003} proposed the hypothesis that there is top-down basic level facilitation from the prefrontal cortex (PFC) during visual object recognition. The top-down signal comes from a "fast" pathway (via fast-responding magnocellular cells) from V2 to the PFC where basic level object categorization is subserved. The signal is then projected back as "initial guesses" to IT, and to be integrated with the bottom-up feed-forward subordinate level object recognition information. More recent work \citep{kveraga2007magnocellular,bar2006top} supports the hypothesis by showing that magnocellular-biased stimuli significantly activated pathways between PFC and IT by increasing the connection strength, based on the human neuroimaging data they collected.

To model the basic level facilitation process based on the development of object categorization, we first train a deep CNN on 308 basic level categories using the ImageNet dataset from the Large Scale Visual Recognition Challenge (ILSVRC) 2012 ~\citep{ImageNetDataset}, and then continue training at the subordinate level on the 1000-way classification task. We show that the top-5 accuracy for 1000-way classification task increases to $82.14\%$, compared to $80.13\%$ achieved by training directly on the subordinate level task using the default parameters in Caffe. We also show that while the facilitation effect tend to appear using many training strategies, the improvement obtained by basic-level pretraining outperforms the others. We then fine-tune this network on Caltech-101~\citep{fei2007learning} and Caltech-256 datasets~\citep{griffin2007caltech}, and show the network trained first on basic level categorization achieves the best generalization. Our result suggests that applying knowledge of human brain on object recognition helps build better models in computer vision tasks.

\section{Method}
Bar's theory of basic level facilitation is consistent with behavioral studies of the development of object categorization that show that young children first learn to categorize basic level objects rather than subordinate or superordinate categories. \citet{mervis1982order} show that children were at ceiling for basic level categories starting from $2\frac{1}{2}$ years of age, but they can only get subordinate level categorization correct until age $5\frac{1}{2}$. \citet{behl1996basic} show that even 3-month-old infants can distinguish pictures of tables from chairs or beds, but they do not display a sensitivity to the differences between furniture and vehicles. A possible reason why basic level categorization is achieved first is that it is most cognitively efficient and the easiest to acquire~\citep{rosch1978principles}. These studies imply that acquiring basic level categorization first may help the development of subordinate level processing.

In this work, we combine the developmental theory and Bar's hypothesis of basic level facilitation for object recognition as a guiding principle for our deep neural network model. We extend Bar's hypothesis to the idea that PFC can train IT at basic level processing before it learns fine-grained distinctions. To implement this idea, we first selected a subset of basic level categories from the ILSVRC 2012 categories and trained a deep CNN on these categories. We then trained the network on a classification task using 1000 categories, starting from the weights learned by the basic level network. The details of how we selected the basic level categories and the following training process are described in the next section.

\subsection{Choosing the Basic Level Categories}
In the prototype theory proposed by \citet{rosch1976basic,rosch1978principles}, basic level categories have the following properties: 1) They share common attributes (e.g., cars) 2)They share the same motor movements (e.g., chairs, a chair is associated with bending of one's knees); 3) They have similar shapes (e.g., apples and bananas); 4) The average shapes of the basic level categories are identifiable. Functionally, basic level categories are thought to be decomposition of the world into maximally informative categories.

We obtained basic level categories from the ImageNet ILSVRC 2012 dataset~\citep{ImageNetDataset}. The dataset contains 1860 object categories (synsets). The synsets are organized using a hierarchical tree, of which 1000 leaf nodes are labeled as the 1000 categories for the classification task in ILSVRC 2012. As there are no explicit labels for the basic level categories, we selected them from all of the 1860 synsets (the 1000 "leaf" categories and the 860 nodes above them). Since the basic level categories are located at various levels of the tree (for example, "dog" has height 5, "fish" is at height 9, and "wolf" is at height 2), we have to find them manually. 
\begin{wrapfigure}{R}{0.4\textwidth}
\begin{center}
\includegraphics[width=0.4\textwidth]{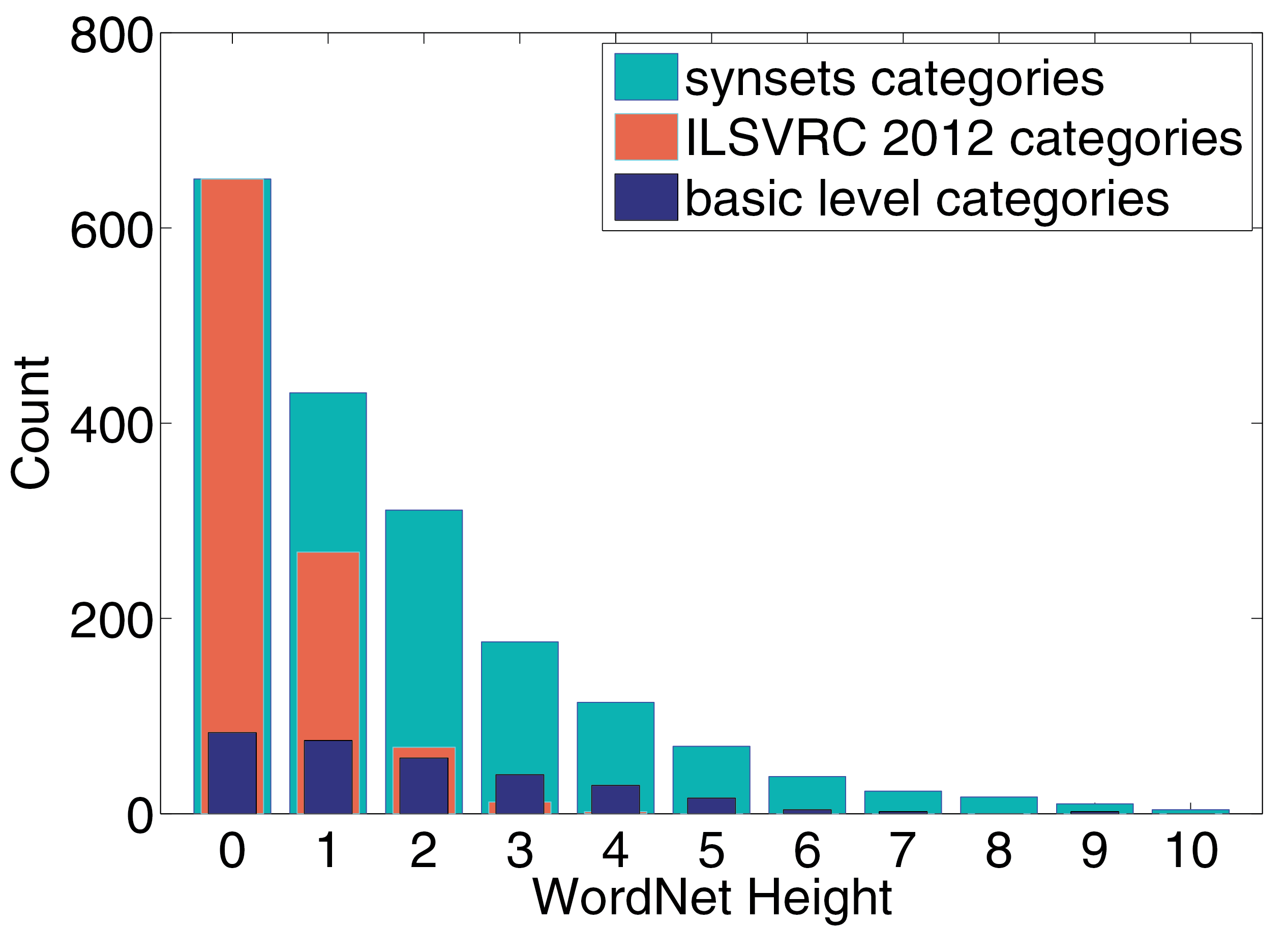}
\end{center}
\caption{Distribution of the basic level categories across height of the ImageNet synsets tree. Most of the categories are located at the lower part of the tree, and the distribution is more balanced than the ILSVRC 2012 categories.}
\label{categorydist}
\end{wrapfigure} 

We did this by using Amazon Mechanical Turk (AMT), where we collected answers of all 1860 synsets for a three-way choice (subordinate, basic, superordinate?) task using the aforementioned properties of basic level categories. We had to arbitrate some disagreements in which were the basic level categories by hand. After obtaining the manually-selected basic level categories, we allocated all descendants in the tree of each category to that category, and assigned a new class label for this new basic level category. If a leaf node belongs to more than one basic level category (for example, "minivan" belongs to "car" and "van"), we simply assigned it to the first ancestor it met. Finally, we obtained a total of 308 basic level categories out of the 1860 synset nodes. Again, not everyone will agree with the final choices, but this will still make the point. Figure~\ref{categorydist} shows the distribution of the basic level categories along with the height of the ImageNet synsets tree. The actual categories we selected are listed in the supplementary material.

After we obtained the 308 basic level classes, we re-labeled all images in the training and validation set of the ILSVRC 2012 dataset. The number of images for each basic level category ranges from to $891$ (hatchet) to $147,873$ (dogs). To reduce the bias of the network towards learning a particular category, we set the maximum number of training images per category to be 4000, which is approximately the mean across all basic level categories. We finally obtained $699,294$ training images to train the basic level network.

Using these 308 basic level categories, we trained a deep convolutional neural network on the basic level categorization task, minimizing the cross-entropy error between the label and network's output:
\[
W_{Basic}^{*}=\mbox{argmin}_{W_{Basic}}\sum_{i}\mathcal{H}(y_{basic}^{(i)},\mbox{ }P_{s}^{(i)}),
\]
where $y_{basic}^{(i)}$ is the label of the basic level category for the $i$th training example, and $P_{s}$ is the
softmax activation of the network. Next, starting from the learned weights $W_{Basic}^{*}$, we trained the network to perform the 1000-way subordinate classification task on the ILSVRC 2012 dataset. We obtained the final weights by optimizing the following function:
\[
W_{Sub}^{*}=\mbox{argmin}_{W_{Sub}}\sum_{i}\mathcal{H}(y_{sub}^{(i)},\mbox{ }P_{s}^{(i)}\mid W_{Basic}^{*}).
\]
\subsection{Relation to Prior Work}
Using basic level facilitation is, in a sense, opposite to the technique of transfer learning. In transfer learning, a large deep network is trained on a large number of categories using a large dataset, such as objects (ImageNet; \citep{ImageNetDataset}) or scenes (Places; \citep{zhou2014learning}). Beginning with the weights learned from the pre-trained network, the network output level is replaced and then just the output weights are trained on datasets of a similar type but with a much smaller number of categories, in order to get better generalization power. 
During transfer learning, the weights of the deep CNN are fixed (except for the last layer), and the transfer learning result is much better than training directly on the smaller dataset~\citep{zeiler2014visualizing}, which often leads to overfitting. 

In contrast, our approach starts by training a network with a relatively small dataset with fewer number of categories than the final dataset to be learned. One can view this approach as a way of doing weight initialization, as it may help to find a good starting point on the error surface of the more complicated task.

More recently, \citet{hinton2015distilling} proposed a curriculum learning training method for deep networks, namely "knowledge distillation (KD)." In KD, a "student" network is not only optimized on the error between the output and the network activation, but also on the error between its own output activation and the (relaxed) output activation of a pre-trained "teacher" network. \citet{hinton2015distilling} show that by adding the knowledge provided by the teacher network, the student network learns better representations.

\citet{romero2014fitnets} extend the idea of KD to hint-based training: the activation of the teacher's hidden layer can serve as hint to a guided hidden layer in the student network using linear regression to add a back-propagated signal from the teacher network. By combining the idea of hint-based training and KD, \citet{romero2014fitnets} show that they can train a thinner but much deeper network more quickly with fewer parameters than the teacher network, with an accompanying boost in generalization accuracy. In our approach, we can think of the "hint" as the weights of the hidden layers of the pre-trained basic level network. We can ultimately extend our model to follow the hint-based learning process using a two-pathway model, which we leave for future work.

\section{Results}
\subsection{Network Training}
The network structure used in this section is exactly the same as AlexNet~\citep{krizhevsky2012} provided in the Caffe deep learning framework~\citep{jia2014caffe}. This method, however, can generally be applied to any network structures and training strategies. The network has 5 hidden convolutional layers and 3 fully connected layers, and the number of feature maps for all layers are $96-256-384-384-256-4096-4096-308$. There are approximately 57 million trainable parameters in the network. We trained our network using stochastic gradient descent with mini-batch size of $256$, momentum of $0.9$, dropout rate of $0.5$, and weight decay of $0.0005$. We set the initial learning rate to $0.01$, and decrease it by a factor of $10$ every $100,000$ iterations. We trained the network for $400,000$ iterations (about $146$ epochs) on a single NVIDIA Titan Black 6GB GPU, which took about 4 days. We achieved a top-5 accuracy of $81.31\%$ on the validation set for the basic level categories.

Starting from the learned basic level network, we continued training the subordinate 1000-way classification task using the whole ILSVRC 2012 dataset. We kept the network structure intact, except for changing the output nodes to 1000 to accommodate the task switch. The 1000-way softmax nodes were initialized using the weights of their corresponding basic level category output weights (for example, the 118 subordinate categories belong to basic category "dog" are initialized using the same trained weights of category "dog" from the pretrained basic level network). In \citet{bar2003}, the fast pathway shares the resources in the early visual cortex (V1 to V2/V4) with the slow pathway. Since the features in V1 to V4 can be characterized by the representation of layer 1 to layer 3 of the deep network~\citep{gucclu2015deep}, we lowered the learning rate of the first 3 convolutional layers to $1/10$ of the higher layers to account for this fact, as they are already learned well. We trained the network for an additional $400,000$ iterations (about $80$ epochs) to make sure the learning converges.

The trained "facilitated" network achieved a top-5 accuracy of $\textbf{82.14\%}$ on the validation set for the 1000-way classification task, comparing to the accuracy of $80.13\%$ using the reference net in Caffe.\footnote{The same result as the benchmark, see: \url{https://github.com/BVLC/caffe/wiki/Models-accuracy-on-ImageNet-2012-val}}. In order to examine whether the improvement is obtained simply because of using more training iterations or fewer categories of pre-training, we performed several additional control experiments: First, we pretrained the reference net for $400,000$ iterations, and continued training the pretrained network for and additional $400,000$ iterations, using the exactly the same training parameters and network structure as the facilitated net. We obtained the top-5 accuracy of $81.16\%$. Second, we pretrained a network using 305 random categories in the ImageNet synset tree that do not overlap with the selected basic-level categories, and trained additional $400,000$ iterations on the 1000-way classification task using the same setting as facilitated network. The top-5 accuracy was $81.17\%$. Third, we initialized the weights to train the facilitated network to be random instead of using the weights from the pretrained basic level network, and the final accuracy was $81.48\%$. The above results suggest that simply having longer training iterations or fewer categories of pre-training is not sufficient to get the improvement that basic-level categorization achieves. Furthermore, the high-level basic-level information (top-layer weights in the CNNs) plays a crucial role to generate this facilitation, which is consistent with Bar's hypothesis. All experimental results are summarized in Table~\ref{result}. 

\begin{table}
\centering{}%
\begin{tabular}{|c||c|}
\hline 
\textbf{Network} & \textbf{Top-5 Accuracy}\tabularnewline
\hline 
\hline 
Reference Net & $80.13\%$\tabularnewline
\hline 
Reference-400K+400K & $81.16\%$\tabularnewline
\hline 
Facilitated-400K+400K & $81.48\%$\tabularnewline
Facilitated-400K+400K \& Basic top layer weights & \textbf{82.14\%}\tabularnewline
\hline 
Random-400K+400K & $81.17\%$\tabularnewline
\hline 
\end{tabular}\caption{Experiment result. A+B means A iterations of pretraining plus B iterations training on ILSVRC 2012 dataset. The network using basic level pretraining and high-level information (top layer weights) of basic level categorization outperforms the others. All networks show some degree of improvement compared to the reference network.}
\label{result}
\end{table}

\subsection{Feature Generalization}
In this section, we explore the generalization power of the learned feature to other datasets, namely Caltech-101 and Caltech-256. We use three models: the basic level pre-trained model (basic), the ImageNet-reference model (reference), and the 1000-way classification model facilitated by the basic level task (facilitated). We keep all except the output layer of our models fixed and train a softmax output layer on top, using the appropriate number of classes of the dataset.

\begin{figure}[t]
\begin{center}
\includegraphics[width=0.99\textwidth]{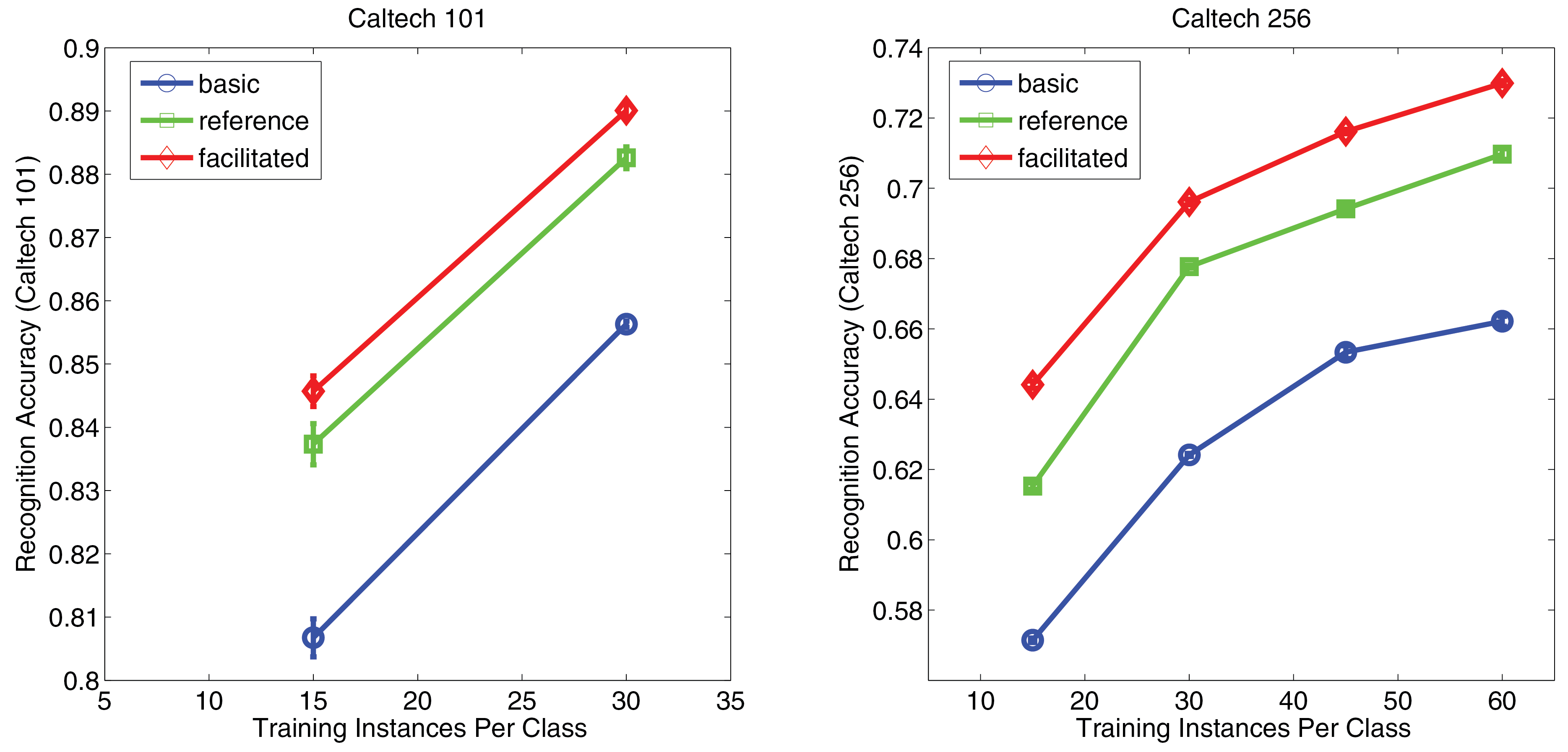}
\end{center}
\caption{Transfer learning results on Caltech-101 (left) and Caltech-256 (right) datasets. We plot the classification accuracy (y axis) as the number of training images per class varies (x axis). Blue line: using the basic level pre-trained model (basic). Green line: using the ImageNet-reference model (reference). Red line: using the 1000-way classification model facilitated by the basic level task (facilitated). Clearly the facilitated model outperforms the other two models.}
\label{caltechresults}
\end{figure}

To avoid contamination of the generalization task due to overlapping images between Caltech datasets and ILSVRC 2012 dataset, we used normalized correlation to identify these "overlap" images, as \citet{zeiler2014visualizing} did. We identified 32 common images (out of 9144 total images) for Caltech-101 dataset and 206 common images (out of 30607 total images) for Caltech-256 dataset, and removed them from the dataset. To evaluate the performance of these datasets, we generated 3 random splits of training data and testing data on these datasets, and computed the averaged performance across the splits. For the Caltech-101 dataset, we randomly selected 15 or 30 images per category to train the output weights, and tested on up to 50 images per class and report the averaged classification accuracy (mean class recall). For the Caltech-256 dataset, we randomly selected 15, 30, 45, or 60 images per category to train the output weights, and tested on up to 50 images per class and report the averaged classification accuracy. The results are reported in Figure~\ref{caltechresults}.

From Figure~\ref{caltechresults}, we can clearly see that the facilitated model performs the best under all conditions. For the Caltech-101 dataset, it achieves the averaged classification accuracy of $89.01\%$ using 30 training examples per class. For the Caltech-256 dataset, it achieves the averaged classification accuracy of $72.99\%$ using 60 training examples per class. The result suggests that the final learned features based on basic level categorization task have better generalization power than training directly from the subordinate level classification task. One thing to note is using the basic level network alone is sufficient to boost the performance to an adequate level (only $3.38\%$ and $6.77\%$ difference to the top performance for Caltech-101 and Caltech-256, respectively), indicating the feature learned by the basic level categorization task alone is already very generic and can be used for other task.

\begin{figure}[ht]
\begin{center}
\includegraphics[width=0.99\textwidth]{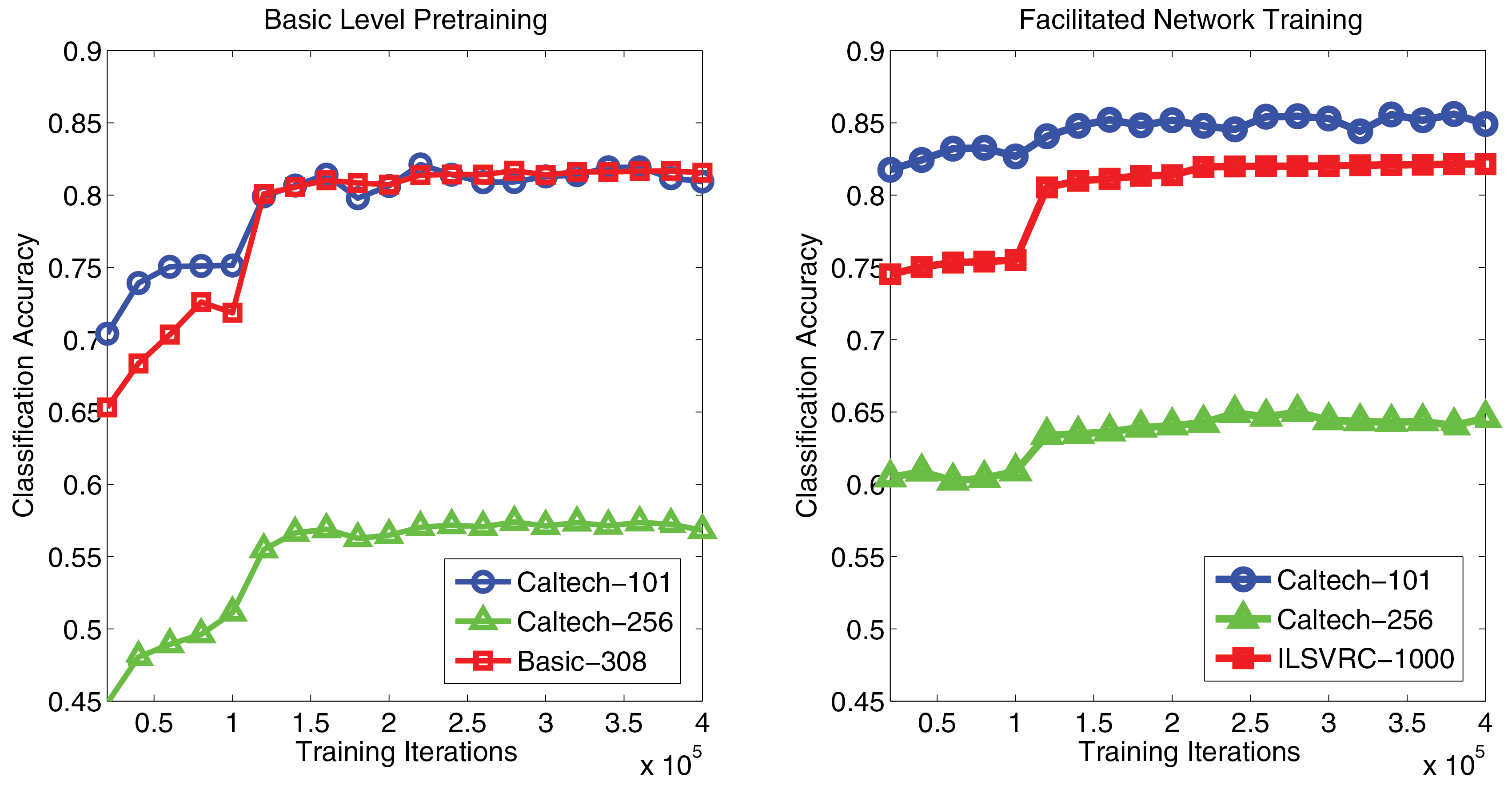}
\end{center}
\caption{Generalization performance of Caltech-101 dataset (blue line) and Caltech-256 dataset (green line) as a function of training iterations in basic level network training (left) and facilitated network training (right). The red line represents the task that the network is training on: basic level categorization (left) and ILSVRC 2012 classification (right).}
\label{caltechresults2}
\end{figure}

In addition, we investigated the learned features through the whole training process to better understand why the basic-level facilitation effect emerges. We did this by measuring transfer to the Caltech-101 and Caltech-256 datasets as a function of training epochs. We examined the curve for simply starting with the basic level network and the facilitated network (i.e., continuing training on 1000 categories). Hence, in the left side of Figure~\ref{caltechresults2}, we start with the basic-level network trained on the 308 basic level categories and then train a new set of output weights for 50K iterations on the Caltech datasets. Hence each point on the red line represents performance on the 308 categories, and the corresponding points on the blue and green lines are the performance on the Caltech datasets after 50K iterations of training, starting with the weights from the red point. The right panel can be thought of as a continuation of the left panel, where the first point corresponds to 20K iterations of training on the 1000-way categorization task\footnote{It would not make sense to start with 0 iterations on the 1000 categories, as the output weights would not be tuned to the categories}. The result is shown in Figure~\ref{caltechresults2}. The boost in performance at 100k iterations in the two graphs are due to lowering the learning rate at that point.

From Figure~\ref{caltechresults2}, we can clearly see that there is a early saturation effect for the Caltech-101 and Caltech-256 datasets, on both basic-level pretraining and facilitated network training. The early saturation effect on basic-level pretraining is easily explained, as the accuracy for 308-way basic-level categorization peaks after 160K training iterations, suggesting the basic-level pretraining can be finished earlier. The early saturation effect on the facilitated network (right panel) is more interesting: although the 1000-way classification accuracy keeps increasing as the training iterations increase, accuracy for Caltech-101 datasets peaks at 160K training iterations and starts fluctuating from then on. This suggests the feature learned for categorizing the more subordinate 1000 ILSVRC 2012 categories may not favor the Caltech-101 dataset, which contains a lot of basic level categories. For the Caltech-256 dataset, however, the peak is later at 280K training iterations. This may be because the Caltech-256 dataset is less biased towards basic-level categories. As the earlier training epochs in the facilitated network may introduce more bias toward the basic-level categorization, this result suggests that better performance on more basic-level biased datasets can be obtained using more basic-level biased feature at earlier stage, and better performance on more subordinate-level biased datasets (like ILSVRC 2012) can be obtained using more subordinate-level biased features at a later stage. The learned feature of basic level categorization, as a guidance, may provides useful information for the subordinate level task. The learned weights by the basic level categorization task (especially the top layer) serve as an excellent starting point on the error surface of the subordinate level task. This information back-projection, or "initial guess", is crucial to help the subordinate level task reach good performance.

\section{Conclusion and Discussion}
We explored the possibility of further optimizing the training of deep networks by adding guiding principles from human development. In particular, we modeled the basic level facilitation effect for visual object recognition, based on data on the development of object categorization~\citep{bornstein2010development} and the basic level facilitation proposed by \citet{bar2003}. We selected basic level object categories from the ImageNet tree hierarchy, trained a basic level categorization network, and continued the training on the 1000-way subordinate level classification task. Our results show that we can get superior classification accuracy using the facilitated network than other training strategies, suggesting the basic level information is a useful prior for the subordinate level classification task. However, the gains are small, and so it is left for future work to assess whether there are better pre-training strategies. 

A more encouraging result is shown in Figure 2, where the network that has been pre-trained on basic level categories shows better transfer to the Caltech datasets than the reference network. Furthermore, this pretraining advantage depends upon training on all 1000 categories after training on basic level ones - there is not good transfer from simply pretraining on basic level categories. This suggests that basic level pretraining is regularizing the network. To the best of our knowledge, this is the first time that the idea of a basic level facilitation effect in visual object recognition has been modeled. 

In our experimental setting, the object category hierarchies are pre-set by ImageNet synset trees, and we selected basic level categories utilizing the tree structure. However, there are methods to automatically find the basic-level categories if the tree is not available, or if the data is unlabeled \citep{marszalek2008constructing,bart2008unsupervised,sivic2008unsupervised}. Although there are other methods to exploit the basic and subordinate level category information\citep{ordonez2015predicting,yan2014hd},our method is a simple CNN structure, and is easily scaled up to more categories.

Our results suggest that we should pay more attention to the structure and neural mechanisms of the visual cortex when building computer vision-related models, especially nowadays, when deep networks are widely used and are considered to be good models of the ventral visual stream. For example, another fact concerning the ventral stream that we have not considered here is that there are two processing pathways: the object recognition pathway through the Lateral Occipital Complex (LOC) and the scene recognition pathway through the PPA. \citet{zhou2014learning} show that by combining features learned in an object recognition network and a scene recognition network, classification results on some datasets improve compared to using a single network. \citet{wang2015bikers} show that combining the information of the entire scene with individual processing helps recognize the urban tribe categories. Clearly, much can be done in this field.

Furthermore, the large number of applications of Artificial Intelligence using deep networks may help us understand more about the brain, especially the visual processes in cortex, such as the development of hemispheric lateralization~\citep{WangCottrell2013}, and the experience moderation effect for object recognition~\citep{WangCottrell2014}. Previous visual processing  models~\citep{Riesenhuber1999,cottrell2011neurocomputational} are shallow and not deep enough to fully characterize the visual pathway. The emergence of deep networks provides us with a more powerful tool to help us  model these cognitive phenomena, thus improving our understanding of the brain.

\subsubsection*{Acknowledgments}
This work was supported in part by NSF Science of Learning Center grants SBE-0542013 and SMA-1041755 to the Temporal Dynamics of Learning Center, and NSF grant IIS-1219252 to GWC. PW was supported by a fellowship from Hewlett-Packard.

\bibliography{iclr2016_conference}
\bibliographystyle{iclr2016_conference}

\newpage
\section*{Supplementary Material: The Basic Level Categories}
Table~\ref{basiclevelcategories} lists all of the 308 basic level categories obtained from the 1860 ILSVRC 2012 synsets. We shorten the names of some categories if they are excessively long.

\begin{center}
\begin{longtable}{lll}
\caption{The Basic Level Categories}\\
\label{basiclevelcategories}\\
abacus                              & fastener, fastening, holdfast, fixing & pole                             \\
acorn                               & fence, fencing                        & pot, flowerpot                   \\
aircraft                            & file, file cabinet, filing cabinet    & power drill                      \\
alcohol, alcoholic drink            & filter                                & prayer rug, prayer mat           \\
altar                               & fish                                  & primate                          \\
ant, emmet, pismire                 & flower                                & printer                          \\
Arabian camel, dromedary            & fly                                   & prison, prison house             \\
arachnid, arachnoid                 & footgear                              & procyonid                        \\
armor, armour                       & footwear                              & promontory, headland, head       \\
artichoke, globe artichoke          & fountain                              & protective garment               \\
ashcan, trash can, garbage can      & four-poster                           & puck, hockey puck                \\
attire, garb, dress                 & fox                                   & quilt, comforter, comfort, puff  \\
baby bed, baby  s bed               & frog, toad, toad frog, anuran         & racket, racquet                  \\
bag                                 & fungus                                & radiator                         \\
Band Aid                            & game equipment                        & radio telescope, radio reflector \\
bannister, banister                 & geyser                                & radio, wireless                  \\
basket, handbasket                  & glass, drinking glass                 & remote control, remote           \\
beacon, lighthouse, beacon light    & robe                                  & ridge                            \\
bear                                & gown                                  & robe                             \\
bee                                 & grille, radiator grille               & rodent, gnawer                   \\
beetle                              & grocery store, grocery                & roof                             \\
bell pepper                         & guillotine                            & rubber eraser, rubber            \\
big cat, cat                        & gymnastic apparatus, exerciser        & salamander                       \\
binder, ring-binder                 & hand blower, blow dryer               & scarf                            \\
bird                                & hand tool                             & scoreboard                       \\
book jacket, dust cover             & hard disc, hard disk, fixed disk      & screen                           \\
bookcase                            & hat, chapeau, lid                     & sea lion                         \\
bovid                               & hatchet                               & seat belt, seatbelt              \\
bowl                                & hay                                   & seat                             \\
box                                 & heater, warmer                        & seed                             \\
brassiere, bra, bandeau             & helmet                                & sewing machine                   \\
bread, breadstuff, staff of life    & hermit crab                           & shaker                           \\
breakwater, groin, groyne           & hip, rose hip, rosehip                & sheath                           \\
breathing device                    & hippopotamus, hippo                   & shelter                          \\
bridge, span                        & homopterous insect, homopteran        & shield                           \\
broom                               & housing, lodging                      & shoji                            \\
bubble                              & hyena, hyaena                         & shop, store                      \\
building, edifice                   & iPod                                  & shore                            \\
bullet train, bullet                & iron, smoothing iron                  & ski                              \\
bus, autobus, coach                 & isopod                                & skirt                            \\
butterfly                           & jean, blue jean, denim                & sled, sledge, sleigh             \\
cabinet                             & jersey, T-shirt, tee shirt            & slot machine, coin machine       \\
camera, photographic camera         & joystick                              & snake, serpent, ophidian         \\
cap, cover                          & keyboard instrument                   & soap dispenser                   \\
headdress, cap                      & keyboard                              & solar dish, solar collector      \\
car mirror                          & kitchen appliance                     & source of illumination           \\
cardoon                             & knife                                 & space bar                        \\
carpenter's kit, tool kit          & lacewing, lacewing fly                & space shuttle                    \\
cash machine, cash dispenser        & lampshade, lamp shade                 & squash                           \\
cassette player                     & lawn mower, mower                     & stage                            \\
cassette                            & leporid, leporid mammal               & stethoscope                      \\
castle                              & lighter, light, igniter, ignitor      & stick                            \\
cat, true cat                       & lizard                                & street sign                      \\
CD player                           & llama                                 & stretcher                        \\
centipede                           & lobster                               & stringed instrument              \\
chain saw, chainsaw                 & loupe, jeweler  s loupe               & suit, suit of clothes            \\
chain                               & lumbermill, sawmill                   & sunglass                         \\
chiffonier, commode                 & magnetic compass                      & support                          \\
cliff, drop, drop-off               & marsupial, pouched mammal             & supporting structure             \\
cloak                               & mashed potato                         & swab, swob, mop                  \\
coelenterate, cnidarian             & mask                                  & sweater, jumper                  \\
coil, spiral, volute, whorl, helix  & maze, labyrinth                       & swimsuit, swimwear               \\
column, pillar                      & measuring cup                         & swine                            \\
comic book                          & measuring instrument                  & switch, electric switch          \\
computer, computing machine         & mechanical device                     & syringe                          \\
condiment                           & memorial, monument                    & table lamp                       \\
cooking utensil, cookware           & menu                                  & tape player                      \\
course                              & military uniform                      & teddy, teddy bear                \\
crab                                & milk can                              & telephone, phone, telephone set  \\
crane                               & mitten                                & television, television system    \\
crayfish, crawfish                  & modem                                 & toilet tissue, toilet paper      \\
crocodilian reptile, crocodilian    & mollusk, mollusc, shellfish           & toiletry, toilet articles        \\
cruciferous vegetable               & monitor                               & top, cover                       \\
cucumber, cuke                      & monotreme, egg-laying mammal          & traffic light, traffic signal    \\
curtain, drape, drapery             & mountain tent                         & trap                             \\
dam, dike, dyke                     & mountain, mount                       & tray                             \\
desk                                & movable barrier                       & triceratops                      \\
diaper, nappy, napkin               & mushroom                              & trilobite                        \\
dictyopterous insect                & musteline mammal, mustelid            & triumphal arch                   \\
dining table, board                 & muzzle                                & turtle                           \\
dish                                & necklace                              & tusker                           \\
disk brake, disc brake              & necktie, tie                          & vacuum, vacuum cleaner           \\
dock, dockage, docking facility     & odonate                               & valley, vale                     \\
dog, domestic dog, Canis familiaris & optical instrument                    & watercraft               \\
doormat, welcome mat                & orthopterous insect, orthopteron      & vessel                           \\
dough                               & oscilloscope, scope                   & viverrine, viverrine mammal      \\
drilling platform, offshore rig     & overgarment, outer garment            & walking stick, walkingstick      \\
dugong, Dugong dugon                & packet                                & wallet, billfold, notecase       \\
ear, spike, capitulum               & paddle, boat paddle                   & wardrobe, closet, press          \\
echinoderm                          & paintbrush                            & weapon, arm                      \\
edentate                            & pajama, pyjama, pj  s, jammies        & weight, free weightt             \\
edible fruit                        & parachute, chute                      & whale                            \\
electric fan, blower                & patio, terrace                        & wheeled vehicle                  \\
electro-acoustic transducer         & pen                                   & whistle                          \\
electronic device                   & pencil sharpener                      & white goods                      \\
elephant                            & percussion instrument                 & wild dog                         \\
entertainment center                & person, individual, someone           & wind instrument, wind            \\
envelope                            & Petri dish                            & window shade                     \\
equine, equid                       & photocopier                           & wing                             \\
espresso                            & pick, plectrum, plectron              & wolf                             \\
fabric, cloth, material, textile    & piggy bank, penny bank                & wooden spoon                     \\
face powder                         & pillow                                & worm                             \\
farm machine                        & plow, plough                          &                                 \\
\end{longtable}
\end{center}

\end{document}